\documentclass{article}
\usepackage{iclr2027_conference,times}
\usepackage{amsmath,amsfonts,bm}

\def\eqref#1{equation~\ref{#1}}
\def\1{\bm{1}}

\DeclareMathAlphabet{\mathsfit}{\encodingdefault}{\sfdefault}{m}{sl}
\SetMathAlphabet{\mathsfit}{bold}{\encodingdefault}{\sfdefault}{bx}{n}

\usepackage{hyperref}
\usepackage{url}
\usepackage{graphicx}
\usepackage{booktabs}
\usepackage{multirow}
\usepackage{array}
\usepackage{microtype}
\usepackage{xcolor}
\usepackage{amsmath,amssymb}
\usepackage{enumitem}
\usepackage{pifont}
\usepackage{algorithm}
\usepackage{algpseudocode}
\usepackage{subcaption}
\usepackage{wrapfig}

\definecolor{paperblue}{HTML}{46647A}
\definecolor{paperteal}{HTML}{3F8F88}
\definecolor{papercoral}{HTML}{C95F54}
\hypersetup{colorlinks=true,linkcolor=paperblue,citecolor=paperteal,urlcolor=paperblue}
\newcommand{\method}{\textsc{MixDetect}}

\title{MixDetect: Word-Level Localization and Quantification of AI Editing}
\author{Hongrui Bao$^{1}$, Yubing Ren$^{2}$, Zhendong Pan$^{3}$,\\
\textbf{Fang Fang$^{2}$, Shi Wang$^{4}$, Yanan Cao$^{2}$}\\[0.6em]
\normalfont\small $^{1}$University of the Chinese Academy of Sciences\\
\normalfont\small $^{2}$Institute of Information Engineering\\
\normalfont\small $^{3}$Shandong University\\
\normalfont\small $^{4}$Visual Information Processing and Learning}
\hypersetup{pdftitle={MixDetect: Word-Level Localization and Quantification of AI Editing},pdfauthor={Hongrui Bao, Yubing Ren, Zhendong Pan, Fang Fang, Shi Wang, Yanan Cao}}

\iclrfinalcopy
\begin{document}
\maketitle
\lhead{}

\begin{abstract}
Large language models are increasingly used to edit human-written text rather than generate entire texts from scratch. Conventional AI-text detectors mainly distinguish human-written from fully AI-generated text, while recent methods for AI-edited text typically provide only a text-level label or editing-degree score. We introduce \method{}, a word-level framework for localizing and quantifying AI editing. \method{} separately predicts whether each word has been edited and, conditional on editing, how substantial the edit is, allowing editing scope and editing intensity to be estimated separately. During training, source--edited pairs are aligned to construct word-level supervision, while inference requires only the input text. Experiments show that \method{} accurately localizes AI-edited words, reflects differences in editing intensity, and reveals different scope--intensity patterns across editing degrees and operations. The overall AI editing magnitude increases under additional AI editing, decreases when AI-generated text is edited by humans, and remains nearly unchanged under ordinary human-to-human editing. The aggregated text-level predictions also perform well on binary and ternary AI-text classification and remain effective under domain and generator shifts. These results show that AI editing can be analyzed beyond a single authorship label or editing-degree score by identifying both where AI editing occurs and how substantial the edits are.
\end{abstract}

\section{Introduction}

The strong text generation capabilities of large language models (LLMs) have made them widely used for editing, polishing, and rewriting human-written text.
The degree of editing can range from minor grammar correction and fluency polishing to paraphrasing and substantial rewriting~\citep{thai2026editlens,saha2025almost}.
Such AI-assisted writing is common in scientific and academic writing~\citep{salvagno2023artificial}, education~\citep{cotton2024chatting}, professional and workplace writing~\citep{noy2023experimental}, and everyday writing tasks~\citep{chatterji2025how}.
Conventional AI-text detectors such as DetectGPT~\citep{mitchell2023detectgpt}, Fast-DetectGPT~\citep{bao2024fastdetectgpt}, Binoculars~\citep{hans2024binoculars}, LastDE++~\citep{xu2025lastde}, and EchoPrompt~\citep{bao2026echoprompt} are designed to distinguish human-written from AI-generated text.
They cannot detect partially AI-edited text or quantify different degrees of AI editing.
For AI-edited text, the task is therefore not simply to determine whether a text is human-written or AI-generated, but to measure the degree of AI editing.

Recent studies have extended AI-text detection to AI-edited and human--AI collaborative text.
Several works formulate the problem as three-class classification over human-written, AI-generated, and AI-edited or mixed text, including MixSet~\citep{zhang2024coauthor}, EditLens~\citep{thai2026editlens}, GPTZero~\citep{adam2026gptzero}, and EVIL-Detect~\citep{bao2026evildetect}.
More recently, RACE~\citep{li2026race} further distinguishes four classes according to the creator and editor of the text.
However, these methods still output a text-level class label or editing score rather than quantify the degree of AI involvement within different parts of the text.

\begin{figure}[t]
\centering
\includegraphics[width=\linewidth]{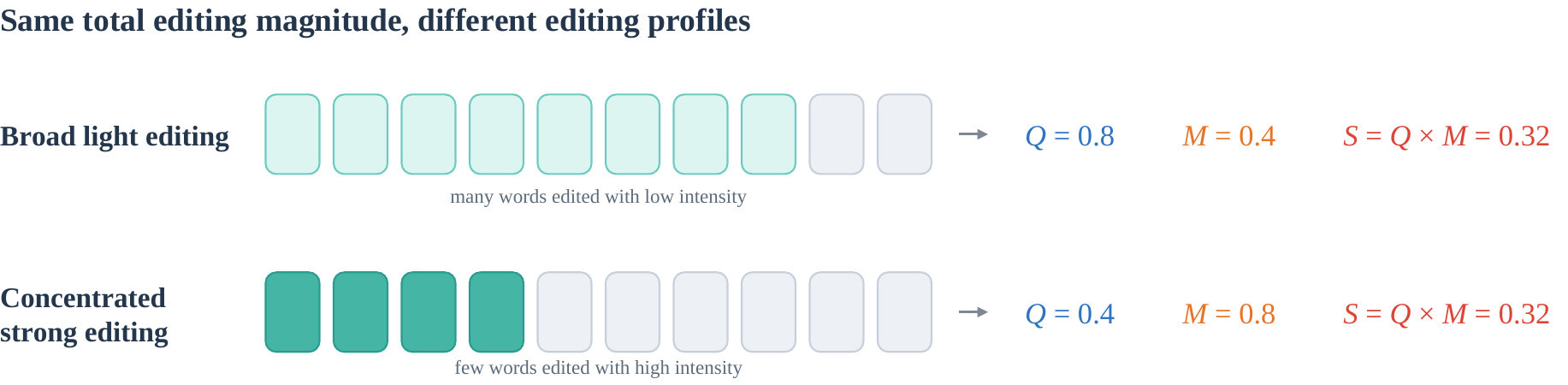}
\caption{
The same overall AI editing magnitude $S$ can arise from different editing patterns.
Broad, light editing has high scope $Q$ but low intensity $M$, whereas concentrated rewriting has lower scope but higher intensity.
Word-level predictions preserve this distinction.
}
\label{fig:task}
\end{figure}

To quantify the degree of AI involvement within a text, we define the \emph{overall AI editing magnitude} as $S$, as illustrated in Figure~\ref{fig:task}.
The overall AI editing magnitude consists of two components: \emph{editing scope} $Q$, which measures the proportion of words affected by AI editing, and \emph{editing intensity} $M$, which measures the average magnitude of the edits among the affected words.
Their product gives the overall AI editing magnitude,
\begin{equation}
S = QM.
\end{equation}
A text with broad but light editing can have a similar $S$ to a text with fewer but more substantial edits.
However, a similar overall AI editing magnitude does not imply the same AI editing behavior, because the two texts can differ substantially in both editing scope and editing intensity.
Therefore, quantifying AI editing requires not only measuring its overall magnitude but also characterizing how the editing differs within the text.

To characterize how AI editing varies within a text, we introduce \textsc{MixDetect}, a word-level framework for localizing and quantifying AI editing.
For each word, \textsc{MixDetect} predicts \emph{edit occurrence}, indicating whether the word is affected by AI editing, and \emph{conditional editing intensity}, estimating how substantial the edit is once editing occurs. Their product defines the corresponding word-level AI-editing score.
These word-level predictions constitute the primary outputs of \textsc{MixDetect}, and, when text-level analysis is needed, they can be aggregated to derive editing scope $Q$, editing intensity $M$, and the overall AI editing magnitude $S$.
We train \textsc{MixDetect} to estimate edit occurrence and intensity using word-level supervision derived from aligned source--edited pairs. Once trained, the model predicts word-level AI-editing scores from the input text, which can be aggregated for text-level analysis or classification.

Experiments show that \textsc{MixDetect} accurately localizes AI-edited words, achieving an AUROC of $0.885$, while its predicted intensity increases with edit magnitude.
The aggregated overall AI editing magnitude is also highly consistent with EditLens, with Spearman correlations of $0.913$ for Edit-Only and $0.939$ for Full.
The overall AI editing magnitude increases after additional AI editing, decreases when AI-generated text is edited by humans, and remains nearly unchanged under ordinary human-to-human editing.
For standard AI-text detection, \textsc{MixDetect} achieves a macro-F1 of $0.916$ on ternary human-written, AI-edited, and AI-generated text classification and remains effective under domain and generator shifts.

Our contributions are threefold:
\begin{itemize}[leftmargin=*,nosep]
\item We formulate the detection of AI editing as a word-level problem that separates \emph{where} editing occurs from \emph{how substantial} the edits are, providing separate estimates of editing scope and intensity rather than only a single overall score.

\item We introduce source-aligned word-level supervision for learning edit occurrence and conditional intensity from source--edited pairs, while requiring only a single input text at inference time.

\item We conduct extensive evaluations across editing degrees, editing operations, and different editing directions, showing that the predictions reflect diverse AI-editing patterns while retaining strong performance on conventional AI-text classification.

\end{itemize}

\section{Related Work}

\noindent\textbf{AI-generated text detection.}
Most AI-text detectors are designed to distinguish human-written from AI-generated text.
Existing methods use token likelihood and rank~\citep{gehrmann2019gltr}, perturbation-based curvature~\citep{mitchell2023detectgpt}, conditional probability curvature~\citep{bao2024fastdetectgpt}, cross-model perplexity ratios~\citep{hans2024binoculars}, local--global probability statistics~\citep{xu2025lastde}, or latent prompt restoration~\citep{bao2026echoprompt}.
These methods typically output a detection score or label for the entire text and are not designed for AI-edited text detection or word-level edit localization.

\noindent\textbf{AI-edited and mixed-authorship text.}
Recent studies extend AI-text detection to AI-edited and mixed human--AI text.
SeqXGPT~\citep{wang2023seqxgpt} and PaLD~\citep{lei2025pald} localize human- and AI-generated regions within mixed-authorship text.
At the text level, EditLens~\citep{thai2026editlens} and EVIL-Detect~\citep{bao2026evildetect} distinguish human-written, AI-generated, and AI-edited text, with EditLens~\citep{thai2026editlens} additionally estimating a continuous editing-degree score.
RACE~\citep{li2026race} further distinguishes four classes according to the creator and editor of the text.
These methods focus on authorship localization or text-level classification and editing-degree estimation rather than quantifying the degree of AI editing within different parts of the text.

\noindent\textbf{AI-editing benchmarks.}
Several recent datasets provide source--edited pairs for studying AI editing.
MixSet~\citep{zhang2024coauthor} includes mixed human--AI text under different editing and co-authoring settings.
EditLens~\citep{thai2026editlens} contains human-written source texts together with AI-edited and fully AI-generated variants, supporting evaluation across different degrees of AI involvement.
APT-Eval~\citep{saha2025almost} varies the requested degree and proportion of AI polishing.
BEEMO~\citep{artemova2025beemo} contains AI responses followed by both model and expert-human edits, while LAMP~\citep{chakrabarty2025lamp} provides AI-generated texts edited by professional writers together with human-annotated edit regions.
These benchmarks cover complementary aspects of AI-assisted editing, including editing degree, editing proportion, editor type, and localized edits.

\section{Method}
\label{sec:method}

\begin{figure}[t]
\centering
\includegraphics[width=\linewidth]{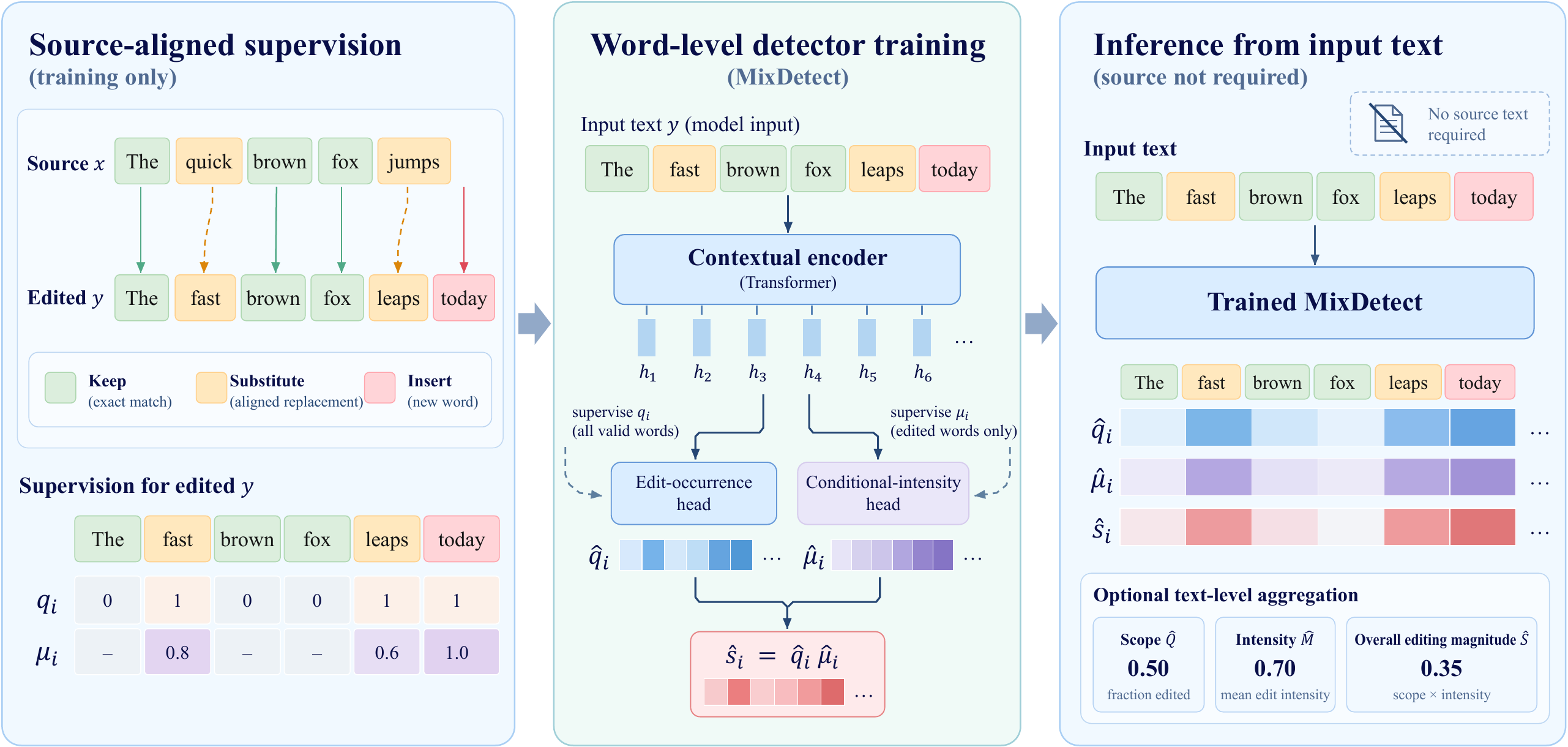}
\caption{Overview of \textsc{MixDetect}. During training, the source--edited pair is aligned to construct word-level supervision for edit occurrence $q_i$ and conditional intensity $\mu_i$. At inference time, the detector receives only the input text and predicts $\hat q_i$ and $\hat\mu_i$ for each word, with $\hat s_i=\hat q_i\hat\mu_i$.}
\label{fig:architecture}
\end{figure}

\noindent\textbf{Overview.}
Figure~\ref{fig:architecture} gives an overview of \textsc{MixDetect}.
\textsc{MixDetect} is designed to detect AI editing at the word level.
For AI-edited training examples, a source--edited pair $(x,y)$ is aligned to construct word-level supervision for edit occurrence $q_i$ and conditional editing intensity $\mu_i$.
The detector takes a single input text and predicts $\hat q_i$ and $\hat\mu_i$ for each word, with $\hat s_i=\hat q_i\hat\mu_i$ representing the corresponding word-level AI-editing score.
These word-level predictions are the primary outputs of \textsc{MixDetect}.
When text-level analysis is required, they can be aggregated to obtain editing scope, editing intensity, and the overall AI editing magnitude.
At inference time, the detector requires only the input text and does not require its source.

\subsection{Source-Aligned Word Supervision}
\label{sec:source_alignment}

Let $x=(u_1,\ldots,u_m)$ denote a human-written source text and $y=(w_1,\ldots,w_n)$ its AI-edited text, where $w_i$ indexes the non-punctuation words considered by the model.
For each word $w_i$, we define an edit-occurrence label $q_i\in\{0,1\}$ indicating whether the word is affected by AI editing.
For an edited word with $q_i=1$, we further define a conditional editing intensity $\mu_i\in[0,1]$ that measures the magnitude of the edit relative to its corresponding source content.
The word-level AI-editing score is then
\begin{equation}
s_i=q_i\mu_i.
\label{eq:word_score}
\end{equation}
Thus, $s_i=0$ for an unchanged word, while larger values indicate stronger edits at edited words.
The conditional intensity $\mu_i$ is undefined when $q_i=0$ and is not supervised for unchanged words.

To construct these labels, we align the source text $x$ with the edited text $y$.
Each word $w_i$ is assigned one of three alignment types: \textsc{Keep} if it is preserved from the source, \textsc{Substitute} if it corresponds to a different source word, and \textsc{Insert} if it has no aligned source word.
AI editing is not strictly word-by-word: words can be inserted, deleted, or replaced, so source and edited words cannot be matched reliably by position alone. Exact lexical matching is also insufficient because an edited word may differ lexically from its source word while preserving similar meaning. We therefore align the source and edited texts in two stages.

First, we identify normalized exact lexical matches between $x$ and $y$ and use contiguous matched spans as anchors. These anchors identify clearly preserved content and divide the remaining unmatched words into smaller intervals. Second, within each interval, we match source and edited words using contextual semantic similarity while preserving their order. If source words $u_j$ and $u_{j'}$ are aligned with edited words $w_i$ and $w_{i'}$, respectively, we require
$$
j<j' \quad \Longrightarrow \quad i<i'.
$$
This constraint prevents crossing matches and reduces spurious correspondences between distant words.

For a source word $u_j$ and an edited word $w_i$, we compute their contextual similarity as
\begin{equation}
c_{ji}=\operatorname{clip}\left(\cos(\mathbf{e}^{x}_j,\mathbf{e}^{y}_i),0,1\right),
\label{eq:word_similarity}
\end{equation}
where $\mathbf{e}^{x}_j$ and $\mathbf{e}^{y}_i$ are their contextual word representations. Within each anchor-delimited interval, dynamic programming finds the order-preserving one-to-one alignment with the largest cumulative similarity. The procedure for obtaining contextual word representations and the dynamic-programming alignment algorithm are provided in Appendix~\ref{app:alignment}.

The resulting alignment directly determines the word-level supervision:
\begin{equation}
(q_i,\mu_i,s_i)=
\begin{cases}
(0,\text{--},0), & w_i\text{ is }\textsc{Keep},\\
(1,1-c_{ji},1-c_{ji}), & w_i\text{ is }\textsc{Substitute},\\
(1,1,1), & w_i\text{ is }\textsc{Insert}.
\end{cases}
\label{eq:word_supervision}
\end{equation}
For a \textsc{Substitute}, $1-c_{ji}$ measures the contextual dissimilarity between the edited word and its aligned source word, so semantically similar substitutions receive lower intensity values and larger changes receive higher values. An \textsc{Insert} has no aligned source word and is assigned the maximum intensity value $\mu_i=1$. For a \textsc{Keep} word, $\mu_i$ is undefined and is not included in the intensity loss.

Punctuation is excluded from supervision. Unmatched source words correspond to deletions; because they have no corresponding position in the edited text, they do not receive word-level labels. The complete alignment and label-construction procedure is given in Algorithm~\ref{alg:source_alignment}.

\subsection{Training the Word-Level Detector}
\label{sec:training}

The source-aligned labels provide an edit-occurrence target $q_i$ for each valid word and a conditional editing-intensity target $\mu_i$ for edited words.
We train a word-level detector to predict both quantities from a single input text.

Given an input text, a contextual encoder produces a word representation $\mathbf{h}_i$ for each word $w_i$.
Two prediction heads operate on the shared representation:
\begin{equation}
\hat q_i
=
\sigma\!\left(
\mathbf{a}_q^{\top}\mathbf{h}_i+b_q
\right),
\qquad
\hat\mu_i
=
\sigma\!\left(
\mathbf{a}_\mu^{\top}\mathbf{h}_i+b_\mu
\right),
\label{eq:prediction_heads}
\end{equation}
where $\hat q_i$ predicts edit occurrence and $\hat\mu_i$ predicts conditional editing intensity.
The predicted word-level AI-editing score is
\begin{equation}
\hat s_i=\hat q_i\hat\mu_i.
\label{eq:predicted_word_score}
\end{equation}

\noindent\textbf{Training labels.}
For human-written texts, every valid word is assigned $(q_i,\mu_i,s_i)=(0,\text{--},0)$. AI-edited texts use the source-aligned labels in Equation~\ref{eq:word_supervision}.

We use \textbf{Edit-Only} as the main setting, trained only on human-written and AI-edited texts. Its purpose is to learn the AI-editing signal relative to human-written text. If the model correctly learns this signal, the predicted scores for fully AI-generated text should also be higher.
We additionally train a \textbf{Full} setting that includes fully AI-generated texts, allowing us to evaluate the effect of these examples on conventional AI-text classification and on fine-grained edit localization and intensity estimation.

Because fully AI-generated texts do not have a corresponding human-written source for source--edited alignment, we assign
\begin{equation}
(q_i,\mu_i,s_i)=(1,1,1)
\label{eq:generated_labels}
\end{equation}
to every valid word in these texts. The two settings use the same model architecture and training objective and differ only in the training data.

\noindent\textbf{Conditional training objective.}
Let $m_i\in\{0,1\}$ indicate whether $w_i$ is a valid non-punctuation word, and define $m_i^{+}=m_iq_i$ so that the intensity loss is applied only to edited words. The model is trained with
\begin{equation}
\mathcal{L}
=
\frac{
\sum_i m_i\,\operatorname{BCE}(\hat q_i,q_i)
+
\sum_i m_i^{+}\,\operatorname{BCE}(\hat\mu_i,\mu_i)
}{
\sum_i m_i
}.
\label{eq:training_objective}
\end{equation}
The occurrence loss is computed for every valid word, whereas the intensity loss is computed only where $q_i=1$. Assigning zero intensity to unchanged words would make the intensity head learn whether an edit occurs in addition to how substantial it is. Masking unchanged words keeps the two prediction targets separate.

Further implementation details are provided in Appendix~\ref{app:implementation}.

\subsection{Inference and Text-Level Aggregation}
\label{sec:inference}

At inference time, \textsc{MixDetect} takes only the input text and does not require its source or the alignment procedure.
For each word $w_i$, the detector outputs edit occurrence $\hat q_i$, conditional editing intensity $\hat\mu_i$, and word-level AI-editing score $\hat s_i=\hat q_i\hat\mu_i$.
These word-level predictions are the primary outputs of \textsc{MixDetect} and are used to localize and characterize AI editing within the text.

For text-level analysis, the word-level predictions are aggregated into editing scope, editing intensity, and overall AI editing magnitude:
\begin{equation}
\hat Q=\frac{1}{n}\sum_{i=1}^{n}\hat q_i,\qquad
\hat M=\frac{\sum_{i=1}^{n}\hat q_i\hat\mu_i}{\sum_{i=1}^{n}\hat q_i},\qquad
\hat S=\frac{1}{n}\sum_{i=1}^{n}\hat q_i\hat\mu_i=\hat Q\hat M.
\label{eq:text_level_aggregation}
\end{equation}
Here, $\hat Q$ denotes the predicted editing scope, $\hat M$ denotes the predicted editing intensity, and $\hat S$ denotes the predicted overall AI editing magnitude.
These text-level predictions are obtained by aggregating the word-level predictions rather than by additional prediction heads.
They are used for text-level editing analysis and evaluation, while the primary outputs of \textsc{MixDetect} remain the word-level predictions.
Appendix~\ref{app:qualitative} shows an example of the predicted word-level editing scores.

\section{Results}
\label{sec:results}

We evaluate \textsc{MixDetect} on both fine-grained AI-editing analysis and standard AI-text detection.
The experiments cover word-level edit localization and intensity estimation,
different editing degrees and operations, changes across editing directions,
as well as text-level classification and ablation studies.
Unless otherwise specified, all \textsc{MixDetect} models are trained on the EditLens training split~\citep{thai2026editlens} and evaluated on each target dataset without further tuning.
Additional details on datasets, implementation, baselines, evaluation metrics, and experimental protocols are provided in Appendix~\ref{app:experimental}.

\subsection{Word-Level Localization and Editing Intensity}
\label{sec:localization_intensity}

We first evaluate whether \textsc{MixDetect} can localize AI-edited words, estimate the intensity of individual edits, and reflect the overall AI editing magnitude through its aggregated word-level predictions consistently with EditLens.

\noindent\textbf{Word-level localization.}
We compare the predicted edit-occurrence score $\hat q_i$ with the word-level labels constructed from source--edited alignment. Table~\ref{tab:localization_factor_recovery} shows that both training settings perform well on edit localization. Edit-Only reaches an AUROC of $0.885$ and an AUPRC of $0.835$, indicating that the detector can accurately localize edited words from the input text.

\begin{table}[t]
\centering
\caption{
Word-level localization and text-level evaluation on the EditLens test split.
AUROC and AUPRC evaluate edit localization over valid words, while AP$_{\mathrm{text}}$ computes average precision within each text and then averages across texts. $\rho_Q$ and $\rho_M$ are partial Spearman correlations for editing scope and intensity, respectively, and $\rho_S$ is the Spearman correlation for overall AI editing magnitude.
}
\label{tab:localization_factor_recovery}
\small
\begin{tabular}{lcccccc}
\toprule
Training
& AUROC
& AUPRC
& AP$_{\mathrm{text}}$
& $\rho_Q$
& $\rho_M$
& $\rho_S$ \\
\midrule
Edit-Only
& 0.885
& \textbf{0.835}
& 0.679
& \textbf{0.878}
& \textbf{0.640}
& \textbf{0.861} \\
Full
& 0.885
& 0.817
& \textbf{0.682}
& 0.873
& 0.617
& 0.852 \\
\bottomrule
\end{tabular}
\end{table}

\noindent\textbf{Editing intensity.}
We next examine whether $\hat\mu_i$ varies with the magnitude of an edit. We divide \textsc{Substitute} words in the EditLens test split into small, medium, and large changes according to their source--edited similarity and report \textsc{Insert} words separately.
\begin{wrapfigure}{l}{0.38\linewidth}
    \centering
    \vspace{-6pt}
    \includegraphics[width=\linewidth]{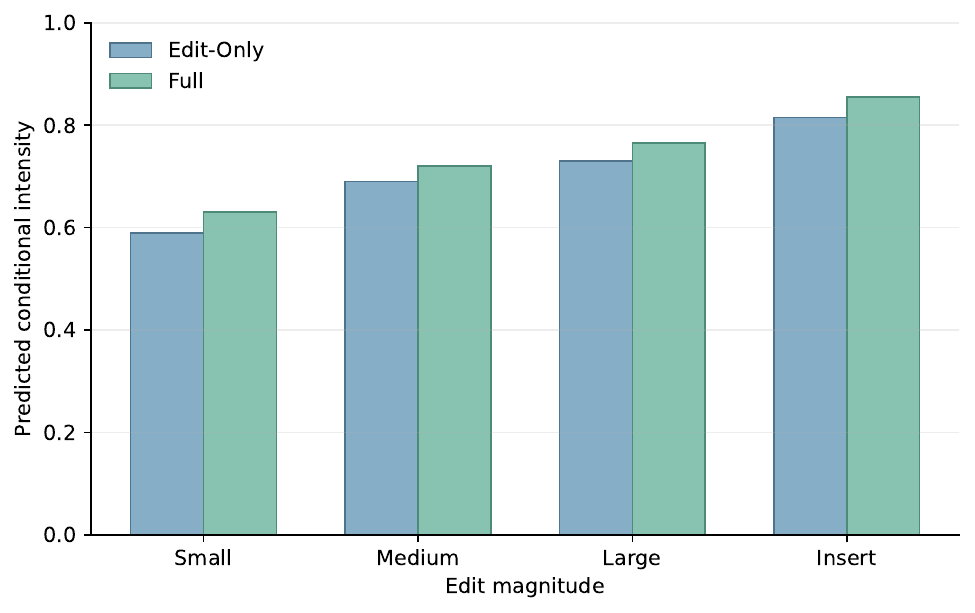}
    \caption{
    Predicted intensity increases with edit magnitude.
    }
    \label{fig:intensity_monotonicity}
    \vspace{-8pt}
\end{wrapfigure}
Figure~\ref{fig:intensity_monotonicity} shows a clear increasing trend for Edit-Only: the mean predicted intensity rises from $0.587$ for small changes to $0.730$ for large changes and reaches $0.817$ for insertions. 
Across substituted words, $\hat\mu_i$ also correlates with the reference intensity ($\rho=0.384$). Full follows the same increasing trend. At the text level, the aggregated predictions also correlate strongly with the corresponding reference values, with Edit-Only reaching $\rho_Q=0.878$, $\rho_M=0.640$, and $\rho_S=0.861$ (Table~\ref{tab:localization_factor_recovery}).

\noindent\textbf{Consistency with EditLens.}
We also compare the overall AI editing magnitude $S$ predicted by \textsc{MixDetect} with the official EditLens editing-degree score on the same test split.
The two measures have Spearman correlations of $0.913$ for Edit-Only and $0.939$ for Full.
Among texts derived from the same source, their pairwise ordering agrees with EditLens in $97.9\%$ of comparisons for Edit-Only and $99.2\%$ for Full.
These results show that the overall AI editing magnitude estimated by \textsc{MixDetect} is highly consistent with EditLens, while \textsc{MixDetect} additionally provides word-level edit localization and editing-intensity estimates.

\subsection{Human-to-AI Editing}
\label{sec:human_ai_editing}

We next evaluate how \textsc{MixDetect} behaves when human-written text undergoes different forms of AI editing. APT-Eval~\citep{saha2025almost} provides different requested editing degrees and proportions, while the Grammarly data~\citep{thai2026editlens} apply different editing operations to the same human sources.

\begin{figure*}[t]
\centering
\includegraphics[width=\textwidth]{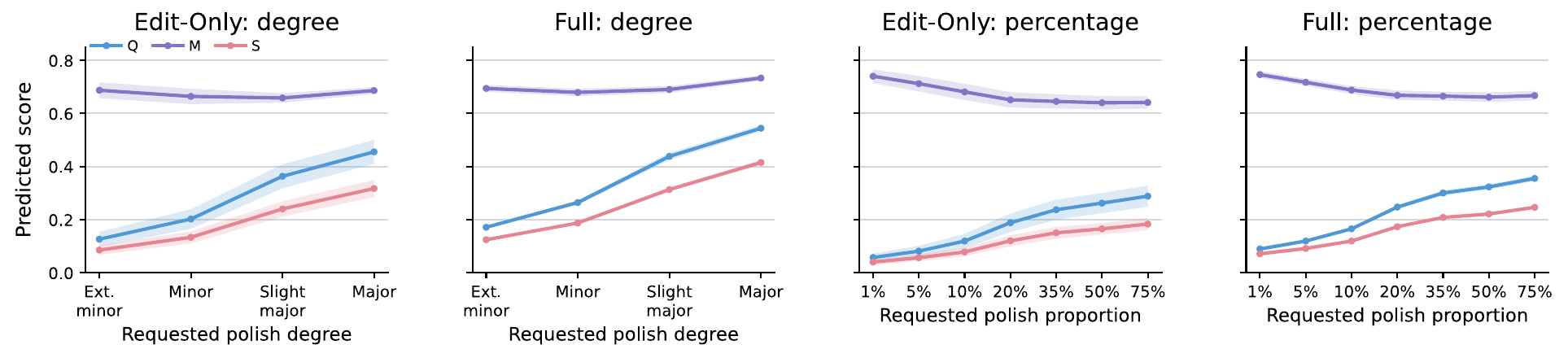}
\caption{
Predicted editing scope, intensity, and overall AI editing magnitude under different requested editing degrees and proportions on APT-Eval.
}
\label{fig:apt_eval_degree_percentage}
\end{figure*}

\noindent\textbf{Editing degree and proportion.}
APT-Eval varies both the requested degree of AI polishing and the requested proportion of the text to be edited. Figure~\ref{fig:apt_eval_degree_percentage} shows that the overall AI editing magnitude $S$ increases with both settings. For Edit-Only, $S$ rises from $0.085$ to $0.317$ across editing degrees, with a Spearman correlation of $0.578$, and from $0.040$ to $0.183$ as the requested editing proportion increases, with a correlation of $0.507$. Full follows the same trends.
The separate scope and intensity predictions show that these changes are mainly reflected in editing scope: $Q$ increases steadily, whereas $M$ remains comparatively stable. This indicates that stronger editing requests primarily affect a larger portion of the text rather than uniformly increasing the magnitude of each individual edit. On the editing-degree setting, Edit-Only also achieves a higher Spearman correlation than EditLens ($0.578$ vs.\ $0.470$).

\begin{figure}[t]
\centering
\includegraphics[width=\linewidth]{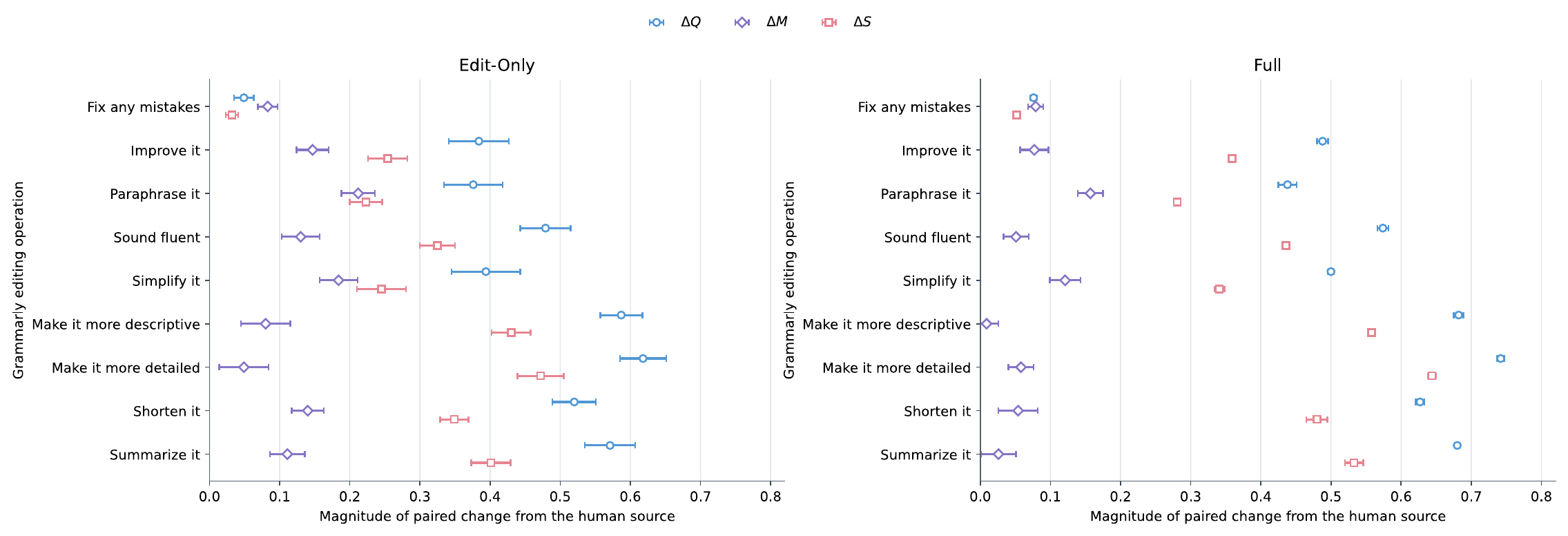}
\caption{
Changes in editing scope, intensity, and overall AI editing magnitude across different Grammarly editing operations. $\Delta Q$, $\Delta M$, and $\Delta S$ denote the magnitudes of the paired score changes relative to the corresponding human source.
}
\label{fig:grammarly_operations}
\end{figure}

\noindent\textbf{Different AI editing operations.}
We further evaluate nine Grammarly editing operations applied independently to the same human sources, covering correction, paraphrasing, rewriting, expansion, shortening, and summarization. Figure~\ref{fig:grammarly_operations} reports the magnitudes of the paired changes in $Q$, $M$, and $S$ relative to the corresponding human source.
The operations produce different scope--intensity patterns. Light correction, such as \textit{Fix any mistakes}, changes all three scores only slightly. Expansion-oriented operations, especially \textit{Make it more detailed} and \textit{Make it more descriptive}, produce the largest changes in $Q$, consistent with their addition of new content across a larger portion of the text. In contrast, \textit{Paraphrase it} produces a relatively large change in $M$, reflecting stronger changes among the affected words without the largest increase in scope. 
The overall AI editing magnitude $S$ correlates with the realized word-level edit distance, with Spearman correlations of $0.719$ for Edit-Only and $0.683$ for Full, and increases after AI editing in $99.5\%$ of the pairs. The different changes in $Q$ and $M$ across operations further show that similar overall AI editing magnitudes can arise from different combinations of scope and intensity.

\subsection{Generalization Across Editing Directions}
\label{sec:direction}

\begin{wraptable}{r}{0.62\linewidth}
    \centering
    \vspace{-6pt}
    \scriptsize
    \setlength{\tabcolsep}{2.5pt}
    \caption{
    Paired changes in the overall AI editing magnitude $S$ on BEEMO under three AI editing prompts (P1--P3).
    }
    \label{tab:beemo_ai_ai}

    \begin{tabular*}{\linewidth}{
        @{\extracolsep{\fill}}llccc@{}
    }
    \toprule
    Detector & Editor
    & $\Delta S$ (P1)
    & $\Delta S$ (P2)
    & $\Delta S$ (P3) \\
    \midrule
    Edit-Only & Llama-3.1-70B
    & +0.097 & +0.150 & +0.158 \\
    Edit-Only & GPT-4o
    & +0.063 & +0.118 & +0.077 \\
    Full & Llama-3.1-70B
    & +0.048 & +0.061 & +0.114 \\
    Full & GPT-4o
    & +0.030 & +0.073 & +0.041 \\
    \bottomrule
    \end{tabular*}

    \vspace{-8pt}
\end{wraptable}
The previous section focuses on human-to-AI editing, which matches the setting used to construct the source-aligned supervision. We next examine how the overall AI editing magnitude changes in the expected direction under three other settings: additional AI editing of AI-generated text, human editing of AI-generated text, and human-to-human editing.

\noindent\textbf{Additional AI editing of AI-generated text.}
Further AI editing of an already AI-generated text should increase $S$ if the score reflects additional AI editing.
BEEMO~\citep{artemova2025beemo} provides this setting, where AI responses are further edited by Llama-3.1-70B~\citep{grattafiori2024llama3} or GPT-4o~\citep{openai2024gpt4o} under three editing prompts.
As shown in Table~\ref{tab:beemo_ai_ai}, $\Delta S>0$ after additional AI editing in all settings.
Thus, $S$ increases under additional AI editing even when the source text is already AI-generated.

\noindent\textbf{Human editing of AI-generated text.}
Human editing of AI-generated text should reduce the overall AI editing magnitude $S$.
BEEMO contains expert-human edits of AI responses, while LAMP~\citep{chakrabarty2025lamp} contains AI-generated texts edited by professional writers together with human-annotated edit regions.
On BEEMO, the average $\Delta S$ is $-0.070$ for Edit-Only and $-0.168$ for Full. $S$ decreases in $66.5\%$ of the pairs for Edit-Only and in $70.4\%$ for Full.
The trend is stronger on LAMP: the average $\Delta S$ is $-0.192$ for Edit-Only and $-0.116$ for Full. $S$ decreases in $96.8\%$ of the pairs for Edit-Only and in $78.0\%$ for Full.
For Edit-Only, larger human edits are associated with larger decreases in $S$. The Spearman correlation between $\Delta S$ and token edit distance is $-0.750$.
The decrease is also concentrated in the edited regions: the local score decreases by $0.309$ inside human-annotated edit regions, compared with $0.106$ in unchanged regions. These local score changes distinguish edited from unchanged regions with an AUROC of $0.769$.
Overall, human editing tends to reduce $S$.

\noindent\textbf{Human-to-human editing.}
Finally, we examine whether ordinary human editing changes $S$.
ArgRewrite~\citep{kashefi2022argrewrite} contains human Draft1--Draft2 editing pairs together with source-matched AI-edited alternatives. For Edit-Only, the mean $S$ scores of the original and human-edited texts are nearly identical ($0.01087$ and $0.01089$, respectively). In contrast, the source-matched AI-edited texts obtain a mean score of $0.594$. In all 86 source families, the AI-edited version receives a higher score than both human versions. Thus, the model does not simply respond to textual change, but distinguishes AI editing from ordinary human editing.

\subsection{Standard AI-Text Classification}
\label{sec:classification}

Although \textsc{MixDetect} is designed for word-level analysis of AI editing, its aggregated text-level predictions can also be used for standard AI-text classification.
We evaluate two binary tasks on the EditLens test split and ternary classification on EditLens, Enron, and Unseen Llama.
The Any AI task distinguishes human-written text from AI-edited or AI-generated text, while Full AI distinguishes fully AI-generated text from human-written and AI-edited text.
The ternary task distinguishes human-written, AI-edited, and fully AI-generated text.
We compare against the official EditLens RoBERTa-large model~\citep{thai2026editlens,liu2019roberta} and four representative AI-text detectors: EchoPrompt~\citep{bao2026echoprompt}, Fast-DetectGPT~\citep{bao2024fastdetectgpt}, Binoculars~\citep{hans2024binoculars}, and LastDE++~\citep{xu2025lastde}.
Qwen2.5-3B Base and Instruct~\citep{yang2024qwen25} are used as proxy models where required by the corresponding detector.
For each method, binary thresholds and ternary threshold pairs are selected on the EditLens validation set to maximize macro-F1 and then fixed for all test and OOD evaluations.

Table~\ref{tab:classification} summarizes the classification results.
On EditLens, Edit-Only achieves the best Any-AI macro-F1 among the compared methods at $0.929$.
Full performs better on tasks that explicitly distinguish fully AI-generated text, reaching $0.980$ on Full AI and $0.916$ on ternary classification.
This difference is consistent with the training data: Full includes fully AI-generated examples, whereas Edit-Only is trained only on human-written and AI-edited text.

\noindent\textbf{Out-of-distribution generalization.}
We further evaluate ternary classification under domain and generator shifts.
Enron~\citep{klimt2004enron} introduces an unseen email domain, while Unseen Llama contains text generated by a model family not observed during training.
As shown in Table~\ref{tab:classification}, Full reaches macro-F1 scores of $0.865$ on Enron and $0.893$ on Unseen Llama, outperforming the compared methods on both datasets.
Edit-Only achieves $0.730$ and $0.828$, respectively, despite not using fully AI-generated examples during training.

\begin{table*}[t]
\centering
\small
\caption{
AI-text classification results on the EditLens test split and under domain and generator shifts.
All binary and ternary results are reported with macro-F1.
}
\label{tab:classification}
\begin{tabular}{lccccc}
\toprule
& \multicolumn{2}{c}{Binary Classification}
& \multicolumn{3}{c}{Ternary Classification} \\
\cmidrule(lr){2-3}
\cmidrule(lr){4-6}
Method
& Any AI
& Full AI
& EditLens
& Enron
& Unseen Llama \\
\midrule
EditLens
& 0.921
& 0.943
& 0.881
& 0.674
& 0.860 \\
EchoPrompt
& 0.722
& 0.768
& 0.607
& 0.660
& 0.432 \\
Fast-DetectGPT
& 0.557
& 0.595
& 0.414
& 0.447
& 0.592 \\
Binoculars
& 0.566
& 0.595
& 0.418
& 0.468
& 0.576 \\
LastDE++
& 0.550
& 0.540
& 0.381
& 0.292
& 0.382 \\
\midrule
\textsc{MixDetect} (Edit-Only)
& \textbf{0.929$\pm$0.002}
& 0.906$\pm$0.012
& 0.848$\pm$0.011
& 0.730
& 0.828 \\
\textsc{MixDetect} (Full)
& 0.928$\pm$0.005
& \textbf{0.980$\pm$0.004}
& \textbf{0.916$\pm$0.005}
& \textbf{0.865}
& \textbf{0.893} \\
\bottomrule
\end{tabular}
\end{table*}

\subsection{Ablation Studies}
\label{sec:ablation}

We examine two design choices of \textsc{MixDetect}: the composition of the training data and the separate prediction of edit occurrence and intensity.

\noindent\textbf{Edit-Only vs.\ Full.}
Edit-Only is trained on human-written and AI-edited text, whereas Full additionally includes fully AI-generated text.
Edit-Only performs better on fine-grained editing analysis, with higher correlations for editing intensity on substituted words ($0.384$ vs.\ $0.348$) and requested editing degree on APT-Eval ($0.578$ vs.\ $0.551$), as well as better localization of human-edited regions on LAMP (AUROC $0.769$ vs.\ $0.646$).
In contrast, Full performs better when fully AI-generated text must also be distinguished, reaching a ternary classification macro-F1 of $0.916$ compared with $0.848$ for Edit-Only and showing stronger OOD classification performance.
Overall, Edit-Only is better suited to fine-grained analysis of partial AI editing, while Full provides better coverage when fully AI-generated text is also considered.

\noindent\textbf{Separate prediction of occurrence and intensity.}
We compare $\hat{s}_i=\hat{q}_i\hat{\mu}_i$ with variants that directly predict a single word-level score, edit occurrence only, or editing intensity only.
Table~\ref{tab:prediction_ablation} shows that Direct-$s$ achieves the best scalar prediction performance, with a word-level AUROC of $0.945$ and a text-level correlation of $0.877$ under Edit-Only, but it cannot distinguish where editing occurs from how substantial the edit is.
The $q$-only model provides edit localization but no intensity estimate, while the $\mu$-only model performs substantially worse in estimating the overall AI editing magnitude because it does not model edit occurrence.
Predicting $q$ and $\mu$ separately therefore provides both editing scope and editing intensity while retaining strong word- and text-level performance.

\begin{table}[t]
\centering
\small
\caption{
Ablation of the prediction formulation. Word AUROC evaluates word-level edit localization, text-level $\rho$ measures the Spearman correlation with the reference overall AI editing magnitude, and Ternary reports macro-F1 for human-written, AI-edited, and AI-generated classification.
}
\label{tab:prediction_ablation}
\begin{tabular}{llcccc}
\toprule
Setting & Prediction
& Word AUROC
& Text-level $\rho$
& Ternary
& Separate $Q,M$ \\
\midrule
\multirow{4}{*}{Edit-Only}
& Direct-$s$
& \textbf{0.945$\pm$0.002}
& \textbf{0.877$\pm$0.005}
& -- & No \\
& $q$ only
& 0.885$\pm$0.005
& 0.822$\pm$0.009
& \textbf{0.850$\pm$0.013}
& No \\
& $\mu$ only
& --
& 0.409$\pm$0.077
& 0.411$\pm$0.011
& No \\
& $q\mu$
& 0.885$\pm$0.005
& 0.861$\pm$0.011
& 0.848$\pm$0.011
& Yes \\
\midrule
\multirow{4}{*}{Full}
& Direct-$s$
& \textbf{0.945$\pm$0.000}
& \textbf{0.880$\pm$0.002}
& -- & No \\
& $q$ only
& 0.885$\pm$0.004
& 0.813$\pm$0.004
& \textbf{0.922$\pm$0.003}
& No \\
& $\mu$ only
& --
& 0.527$\pm$0.006
& 0.652$\pm$0.012
& No \\
& $q\mu$
& 0.885$\pm$0.004
& 0.852$\pm$0.009
& 0.916$\pm$0.005
& Yes \\
\bottomrule
\end{tabular}
\end{table}

\section{Conclusion}

We introduced \textsc{MixDetect}, a word-level framework for localizing and quantifying AI editing.
For each word, \textsc{MixDetect} predicts edit occurrence and conditional editing intensity, providing both localized AI-editing scores and, when needed, aggregated text-level predictions of editing scope, intensity, and overall AI editing magnitude.
Source--edited pairs are used only to construct word-level supervision during training, while inference requires only a single input text.
Experiments show that the model accurately localizes AI-edited words, reflects differences in editing intensity, and produces meaningful changes in scope and intensity across editing degrees, operations, and editing directions; its aggregated predictions also remain effective for standard AI-text classification under domain and generator shifts.
Overall, \textsc{MixDetect} provides a finer-grained view of AI-edited text by showing not only how much AI editing is present, but also where it occurs and how substantial the edits are.

\section*{Ethics Statement}

This work studies the detection and characterization of AI-edited text. Our experiments use existing research datasets and publicly available language models, and do not involve the collection of new personal data or human-subject experiments. The proposed method estimates evidence of AI editing from textual patterns and does not establish the authorship or provenance of a text with certainty. Its predictions may also vary across domains, writing styles, languages, and generation or editing models not represented in the evaluation data. Therefore, \textsc{MixDetect} should not be used as the sole basis for high-stakes decisions about authorship, academic misconduct, or other individual-level judgments. We view fine-grained editing estimates as supporting evidence for analysis rather than definitive attribution.

\section*{Reproducibility Statement}

The source-aligned supervision, word-level prediction heads, and training objective are specified in Sections~\ref{sec:source_alignment}--\ref{sec:training}, while inference and text-level aggregation are described in Section~\ref{sec:inference}. Appendix~\ref{app:alignment} gives the complete alignment and label-construction procedure, including contextual word encoding and the monotonic dynamic program. Appendix~\ref{app:experimental} documents the model configuration, dataset roles, baselines, evaluation metrics, and classification threshold selection used in the experiments. All reported \textsc{MixDetect} results use fixed dataset splits and are averaged over the stated random seeds; validation-selected classification thresholds are frozen before test and out-of-distribution evaluation.

\section*{AI Use Statement}

In this work, we used generative AI tools to assist with method implementation, translation, dataset cleaning and reformatting, scientific figure creation, and editing the manuscript for readability. We did not use generative AI tools for other research tasks requiring disclosure under the ICLR 2027 policy, including developing research hypotheses or methodology, formulating theoretical or mathematical claims, or interpreting experimental results. All AI-assisted code, data processing, translations, figures, and manuscript edits were manually reviewed and verified against the intended methods, original data, and experimental results. All reported results were obtained from the verified implementation and processed data. We take full responsibility for the final content of this work.

\bibliography{references}
\bibliographystyle{iclr2027_conference}

\appendix
\newpage
\section{Source--Edited Alignment Details}
\label{app:alignment}

This section provides the implementation details of the source--edited alignment used to construct the word-level supervision in Section~\ref{sec:source_alignment}.
The procedure first identifies exact lexical anchors and then applies monotonic semantic matching within the unmatched intervals between anchors.

\subsection{Word Preprocessing and Contextual Word Representations}

We first segment the source and edited texts into words while retaining their character spans and punctuation information.
For lexical matching, each non-punctuation word is normalized by lowercasing and removing boundary punctuation:
\begin{equation}
\nu(w)
=
\operatorname{StripBoundaryPunct}
\bigl(\operatorname{Lowercase}(w)\bigr).
\end{equation}

To compute semantic similarity, we obtain contextual word representations for the source and edited texts independently using the same contextual encoder.
Because a word may be split into multiple subword tokens, its contextual word representation is obtained by mean-pooling the hidden states of all subword tokens corresponding to that word:
\begin{equation}
\widetilde{\mathbf e}_i
=
\frac{1}{|\mathcal T(i)|}
\sum_{k\in\mathcal T(i)}
\mathbf z_k.
\end{equation}
Here, $\mathcal T(i)$ denotes the set of subword tokens corresponding to word $i$, and $\mathbf z_k$ is the hidden state of subword token $k$.

For texts longer than the encoder context window, we encode the text using overlapping windows.
If a word appears in multiple windows, its representations are averaged before computing the final normalized word representation:
\begin{equation}
\mathbf e_i
=
\frac{
\frac{1}{|\mathcal W(i)|}
\sum_{r\in\mathcal W(i)}
\widetilde{\mathbf e}^{(r)}_i
}{
\left\|
\frac{1}{|\mathcal W(i)|}
\sum_{r\in\mathcal W(i)}
\widetilde{\mathbf e}^{(r)}_i
\right\|_2
}.
\end{equation}
Here, $\mathcal W(i)$ contains the windows covering word $i$.
The resulting $L_2$-normalized word representations are used to compute the cosine similarities in Equation~\ref{eq:word_similarity}.

\subsection{Exact Lexical Anchors}

We identify contiguous source--edited spans whose normalized word sequences match exactly. These matched spans are used as lexical anchors and are assigned \textsc{Keep}. The anchors partition the remaining source and edited words into independent unmatched intervals. Semantic alignment is then performed separately within each interval rather than globally over the complete text.

\subsection{Monotonic Semantic Alignment}

Consider an unmatched source interval $U=(u_1,\ldots,u_r)$ and edited interval $V=(w_1,\ldots,w_c)$. For each pair $(u_i,w_j)$, we compute the compatibility
\begin{equation}
    A_{ij}
    =
    \operatorname{clip}
    \left(
        \cos(\mathbf e^x_i,\mathbf e^y_j),
        0,1
    \right).
\end{equation}
We then find the maximum-weight one-to-one matching subject to the monotonicity constraint that matched word order is preserved.

Let $D_{i,j}$ denote the maximum cumulative similarity obtainable from the prefixes $(u_1,\ldots,u_i)$ and $(w_1,\ldots,w_j)$. The dynamic program is
\begin{equation}
D_{i,j}
=
\max
\left\{
    D_{i-1,j},
    D_{i,j-1},
    D_{i-1,j-1}+A_{ij}
\right\},
\label{eq:alignment_dp}
\end{equation}
with $D_{0,j}=D_{i,0}=0$. The first two transitions leave a source or edited word unmatched, respectively, while the third aligns $u_i$ with $w_j$. The complete dynamic-programming procedure is summarized in Algorithm~\ref{alg:monotonic_alignment}.

\begin{algorithm}[t]
\caption{Monotonic Semantic Alignment}
\label{alg:monotonic_alignment}
\begin{algorithmic}[1]
\Require Source interval $U=(u_1,\ldots,u_r)$;
         edited interval $V=(w_1,\ldots,w_c)$;
         compatibility matrix $A$
\Ensure Monotonic one-to-one matching $\mathcal M$

\State Initialize $D_{0,j}=D_{i,0}=0$

\For{$i=1,\ldots,r$}
    \For{$j=1,\ldots,c$}
        \State
        $D_{i,j}\gets
        \max\{
            D_{i-1,j},
            D_{i,j-1},
            D_{i-1,j-1}+A_{ij}
        \}$
        \State Store the corresponding backpointer
    \EndFor
\EndFor

\State Backtrack from $(r,c)$ to recover matched word pairs
\State Break score ties by preferring more matched pairs, followed by a fixed
deterministic operation priority
\State \Return $\mathcal M$
\end{algorithmic}
\end{algorithm}

Matched non-identical pairs are treated as \textsc{Substitute}, unmatched edited words as \textsc{Insert}, and unmatched source words as deletions. The latter have no position in the edited text and therefore do not receive word-level supervision.

\subsection{Complete Label-Construction Procedure}

Algorithm~\ref{alg:source_alignment} summarizes the complete supervision construction used in the main text.

\begin{algorithm}[t]
\caption{Source-Aligned Word Supervision}
\label{alg:source_alignment}
\begin{algorithmic}[1]
\Require Source text $x=(u_1,\ldots,u_m)$; edited text $y=(w_1,\ldots,w_n)$
\Ensure Word-level targets $\{q_i,\mu_i,s_i\}_{i=1}^{n}$
\State Segment $x$ and $y$ into words and identify punctuation
\State Normalize non-punctuation words for lexical matching
\State Encode $x$ and $y$ to obtain contextual word representations
\State Find contiguous exact lexical matches and use them as anchor blocks
\For{each edited word $w_i$ in an exact anchor}
    \State Assign $w_i\gets\textsc{Keep}$, $q_i\gets0$, $\mu_i\gets\bot$, $s_i\gets0$
\EndFor
\State Partition the unmatched source and edited words into anchor-delimited intervals
\For{each unmatched source--edited interval $(U,V)$}
    \State Compute pairwise contextual similarities $c_{ji}$
    \State $\mathcal M\gets\Call{MonotonicAlign}{U,V,\{c_{ji}\}}$
    \For{each matched pair $(u_j,w_i)\in\mathcal M$}
        \State Assign $w_i\gets\textsc{Substitute}$, $q_i\gets1$, $\mu_i\gets1-c_{ji}$, $s_i\gets1-c_{ji}$
    \EndFor
    \For{each unmatched edited word $w_i\in V$}
        \State Assign $w_i\gets\textsc{Insert}$, $q_i\gets1$, $\mu_i\gets1$, $s_i\gets1$
    \EndFor
\EndFor
\State Exclude punctuation positions from supervision
\State Record unmatched source words as deletions without word-level labels
\State \Return $\{q_i,\mu_i,s_i\}_{i=1}^{n}$
\end{algorithmic}
\end{algorithm}

\section{Qualitative Example}
\label{app:qualitative}

\begin{figure}[t]
    \centering
    \includegraphics[width=\linewidth]{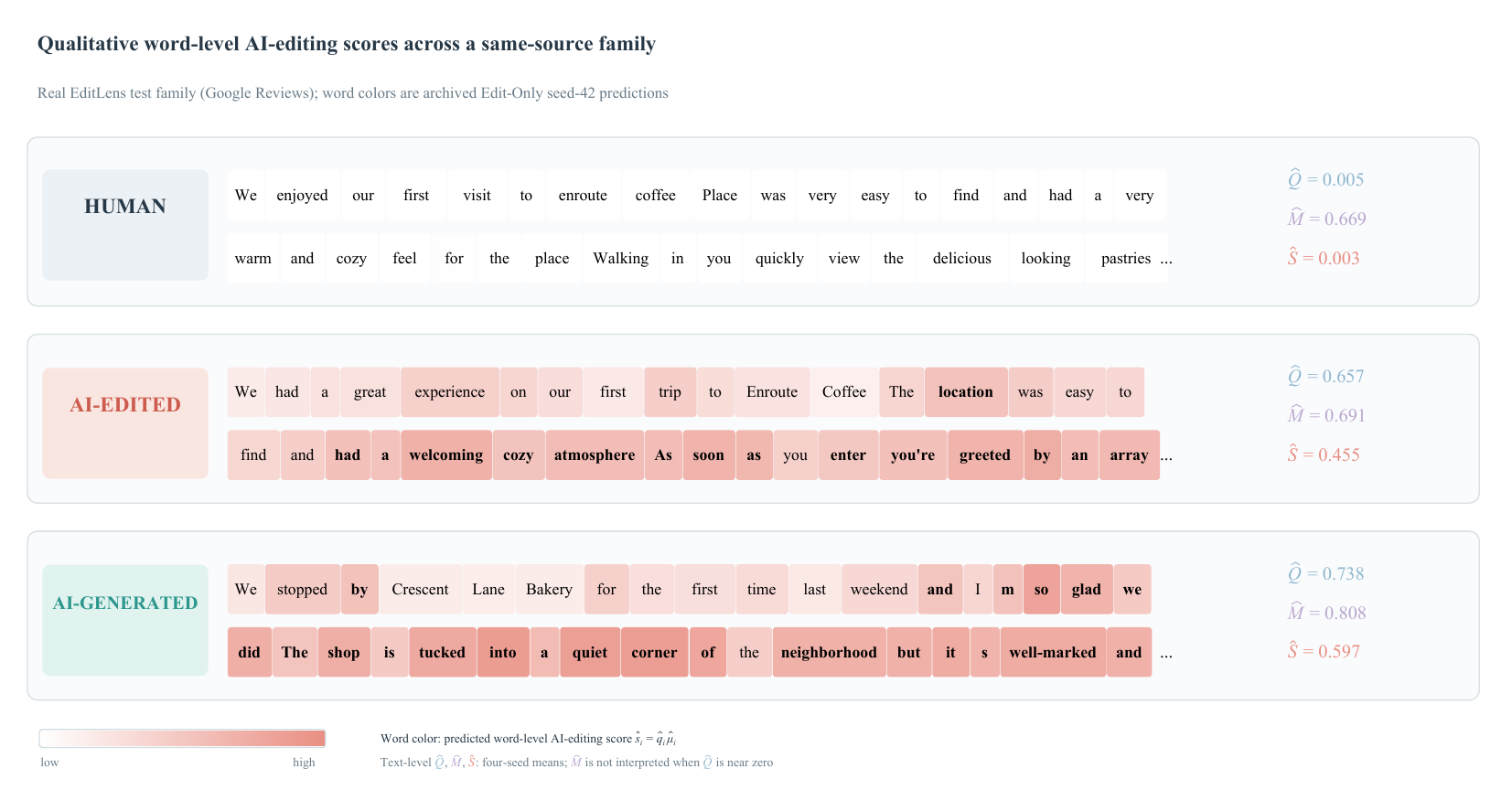}
    \caption{
        Qualitative word-level example from the EditLens test set.
        Word colors indicate the predicted word-level AI-editing score $\hat s_i=\hat q_i\hat\mu_i$, with text-level $\hat Q$, $\hat M$, and $\hat S$ shown on the right.
}
    \label{fig:qualitative_same_source}
\end{figure}

Figure~\ref{fig:qualitative_same_source} shows a same-source example from the EditLens test set, illustrating the word-level AI-editing scores predicted by \textsc{MixDetect} for human-written, AI-edited, and fully AI-generated text.

\section{Experimental Details}
\label{app:experimental}

\subsection{Implementation Details}
\label{app:implementation}

We implement \textsc{MixDetect} using DeBERTa-v3-base~\citep{he2023debertav3} as the shared contextual encoder. When a word is split into multiple subword tokens, we mean-pool their final-layer hidden states to obtain its word representation $\mathbf{h}_i$. The occurrence and intensity predictors are implemented as two independent linear layers followed by sigmoid activations. Edit-Only and Full use the same model architecture and optimization settings and differ only in their training data.

We train the model with AdamW~\citep{loshchilov2019adamw} for three epochs using a learning rate of $2\times10^{-5}$ and a batch size of $8$. The maximum input length is $512$ tokens. Texts exceeding this length are processed with overlapping windows using a stride of $64$. When a word appears in multiple windows, its predictions are averaged before computing the word-level outputs and text-level predictions.

All main \textsc{MixDetect} results are reported as the mean and standard deviation over four random seeds (42--45). The Direct-$s$ ablation uses the same encoder, data, optimization settings, and training schedule, replacing the two prediction heads with a single sigmoid head trained directly on $s_i$; its reported results use seeds 42--44.

\subsection{Datasets and Evaluation Settings}
\label{app:datasets}

\noindent\textbf{EditLens.}
The primary training, validation, and in-distribution test data come from the EditLens dataset~\citep{thai2026editlens}, which contains human-written source texts, AI-edited variants, and fully AI-generated variants. Edit-Only is trained on the human-written and AI-edited examples, while Full additionally includes the fully AI-generated examples described in Section~\ref{sec:training}. The official EditLens training split is used for model fitting, its validation split is used only for threshold selection in text-level classification, and its test split is used for the main localization, text-level evaluation, and classification results. The \texttt{test\_enron} and \texttt{test\_llama} splits released with EditLens are used for domain- and generator-shift evaluation, respectively.

\noindent\textbf{APT-Eval.}
APT-Eval~\citep{saha2025almost} provides human-written texts polished by multiple LLMs under requested editing degrees and requested editing proportions. We evaluate the frozen EditLens-trained \textsc{MixDetect} checkpoints directly on these examples without further tuning. For degree-based evaluation, Spearman correlation is computed between the ordered requested degree and each text-level score. For percentage-based evaluation, the requested editing proportion is treated as an ordered variable and the corresponding score trajectories are reported.

\noindent\textbf{Grammarly.}
The Grammarly evaluation follows the editing-operation setting released with EditLens~\citep{thai2026editlens}. It contains nine independent Grammarly editing operations applied to the same human sources; the operations are evaluated independently rather than as sequential multi-pass edits. For each source--edited pair, we compute the paired changes in $Q$, $M$, and $S$ and compare $S$ with the realized word-level edit distance.

\noindent\textbf{BEEMO.}
BEEMO~\citep{artemova2025beemo} contains AI responses followed by both LLM-based and expert-human edits. For AI-to-AI evaluation, each base response is paired with edits produced by Llama-3.1-70B-Instruct~\citep{grattafiori2024llama3} or GPT-4o~\citep{openai2024gpt4o} under the three BEEMO editing prompts P1--P3. For AI-to-human evaluation, the same base responses are paired with the corresponding expert-human edits. All reported changes are paired differences relative to the underlying AI base response.

\noindent\textbf{LAMP.}
LAMP~\citep{chakrabarty2025lamp} contains AI-generated passages edited by professional writers and provides human-annotated edit regions. We compare text-level score changes before and after professional editing. For regional evaluation, annotated edit spans are mapped to model words and the score reduction inside those spans is compared with the reduction in background words from the same texts.

\noindent\textbf{ArgRewrite.}
ArgRewrite~\citep{kashefi2022argrewrite} provides human Draft1--Draft2 editing pairs. We use these pairs as a human-to-human editing control and compare them with source-matched AI-edited alternatives in the evaluation construction. The reported 86 source families are those for which the original, human-edited, and source-matched AI-edited versions are all available.

\noindent\textbf{Enron and unseen Llama.}
The Enron evaluation uses the email-domain split derived from the Enron corpus~\citep{klimt2004enron} and released as \texttt{test\_enron} with EditLens. The unseen-generator evaluation uses the EditLens \texttt{test\_llama} split, whose generated texts come from a Llama model family~\citep{grattafiori2024llama3} not observed during training. No additional tuning is performed on either split.

\subsection{Baselines}
\label{app:baselines}

We compare against the official EditLens~\citep{thai2026editlens} RoBERTa-large~\citep{liu2019roberta} model and four representative AI-generated-text detectors: EchoPrompt~\citep{bao2026echoprompt}, Fast-DetectGPT~\citep{bao2024fastdetectgpt}, Binoculars~\citep{hans2024binoculars}, and LastDE++~\citep{xu2025lastde}. The zero-shot detectors use Qwen2.5-3B Base and Instruct models~\citep{yang2024qwen25} as the proxy models where their scoring rules require base or instruction-tuned language models. Each baseline retains its own tokenizer and scoring rule, and score directions are fixed before test evaluation. We do not convert text-level baseline scores into artificial word-level scores for the localization experiments.

EchoPrompt requires the reconstructed prompt and response to fit within the model context window. When the prompt would be truncated, we exclude that example because the resulting score no longer corresponds to the complete prompt--response construction. Under this criterion, EchoPrompt covers $53.3\%$ of the EditLens test set, $92.2\%$ of Enron, and $56.5\%$ of unseen Llama. Its reported results are computed over these valid subsets.

\subsection{Evaluation Metrics}
\label{app:metrics}

Word-level AUROC and AUPRC pool all valid non-punctuation words.
AP$_{\mathrm{text}}$ computes average precision independently within each text
and then averages across texts, preventing longer texts from dominating the metric.
For text-level evaluation, $\rho_Q$ and $\rho_M$ are partial Spearman correlations
with the alignment-derived $Q$ and $M$, respectively, while controlling for the
other quantity. $\rho_S$ is the Spearman correlation between the predicted and
reference overall AI editing magnitude.

For paired editing-direction experiments, we report the mean paired score change and the fraction of pairs moving in the expected direction. APT-Eval additionally reports Spearman correlation with ordered requested editing degree or proportion. For LAMP regional evaluation, word-level score reduction is used to distinguish human-annotated edited regions from background words, reported with AUROC (and, where applicable, AUPRC).

\subsection{Classification Protocol}
\label{app:classification_protocol}

We evaluate two binary tasks and one ternary task using the text-level scores. The \emph{Any AI} task distinguishes human-written text from the union of AI-edited and fully AI-generated text. The \emph{Full AI} task distinguishes fully AI-generated text from the union of human-written and AI-edited text. For each detector and each binary task, a decision threshold is selected on the EditLens validation split to maximize macro-F1.

For ternary classification, two ordered thresholds partition the scalar score into human-written, AI-edited, and fully AI-generated regions. The threshold pair is selected on the EditLens validation split to maximize ternary macro-F1. All selected thresholds are then frozen and applied unchanged to the EditLens test split, Enron, and unseen Llama evaluations.

\end{document}